\documentclass[preprint,12pt]{elsarticle}

\usepackage{amssymb}
\usepackage{amsmath}

\begin{document}

\begin{frontmatter}

\title{Sustainable and Intelligent Public Facility Failure Management System Based on Large Language Models}

\author[1]{Siguo Bi}
\author[1]{Jilong Zhang\corref{cor1}}
\ead{jlzhfd@fudan.edu.cn}
\author[2]{Wei Ni}

\cortext[cor1]{Corresponding author}
\affiliation[1]{organization={Library, Fudan University},
	city={Shanghai},
	postcode={200433},
	country={China}}

\affiliation[2]{organization={CSIRO \& UNSW},
	city={Sydney},
	postcode={NSW 2122},
	country={Australia}}

\begin{abstract}
This paper presents a new Large Language Model (LLM)-based Smart Device Management framework, a pioneering approach designed to address the intricate challenges of managing intelligent devices within public facilities, with a particular emphasis on applications to libraries. Our framework leverages state-of-the-art LLMs to analyze and predict device failures, thereby enhancing operational efficiency and reliability. Through prototype validation in real-world library settings, we demonstrate the framework's practical applicability and its capacity to significantly reduce budgetary constraints on public facilities. The advanced and innovative nature of our model is evident from its successful implementation in prototype testing. We plan to extend the framework's scope to include a wider array of public facilities and to integrate it with cutting-edge cybersecurity technologies, such as Internet of Things (IoT) security and machine learning algorithms for threat detection and response. This will result in a comprehensive and proactive maintenance system that not only bolsters the security of intelligent devices but also utilizes machine learning for automated analysis and real-time threat mitigation. By incorporating these advanced cybersecurity elements, our framework will be well-positioned to tackle the dynamic challenges of modern public infrastructure, ensuring robust protection against potential threats and enabling facilities to anticipate and prevent failures, leading to substantial cost savings and enhanced service quality.
\end{abstract}

%

\begin{keyword}
	Public Facility \sep Smart Devices \sep Large Language Models \sep Sustainability \sep Failure Management 



\end{keyword}

\end{frontmatter}



\section{Introduction}
Large Language Models (LLMs) are gaining increasing prominence and importance in international applications. They have demonstrated the potential to enhance governance efficiency in government affairs and are showing strong application prospects in education, healthcare, finance, and industry, among other sectors \cite{Truong2024658,Mekrache202429,Jin2024,Arora2023641}. LLMs are pushing the boundaries of artificial intelligence by integrating multimodal data, exerting a profound impact on society \cite{Sandmann2024}. With their robust capabilities in information integration, understanding, and production, these models are emerging as versatile personal information assistants, accelerating scientific discoveries, and fostering new forms of content creation in the creative industries \cite{Patsakis2023}. Therefore, the application of LLMs across various fields is crucial for improving efficiency, fostering innovation, and strengthening services.

Amidst the ongoing trends of economic globalization and urbanization, urban development has shifted its focus towards sustainable growth, with public facilities such as libraries and museums playing a pivotal role \cite{tan2020cultural,Bi2022584,Bi2022}. In 2023, the Institute of Museum and Library Services (IMLS) allocated a substantial \$266.4 million to projects aimed at enhancing the capabilities of museums, libraries, and associated institutions, underscoring the importance of research and optimization in the substantial investment in intelligent devices within these public facilities \cite{imls_website}. These investments not only elevate the operational efficiency of public facilities but are also crucial to the city's cultural capital and socio-economic development \cite{oteros2018using,rivero2020sustainable,imls_2021_afr}.

\begin{figure}[h]
	\centering
	\includegraphics[width=\linewidth]{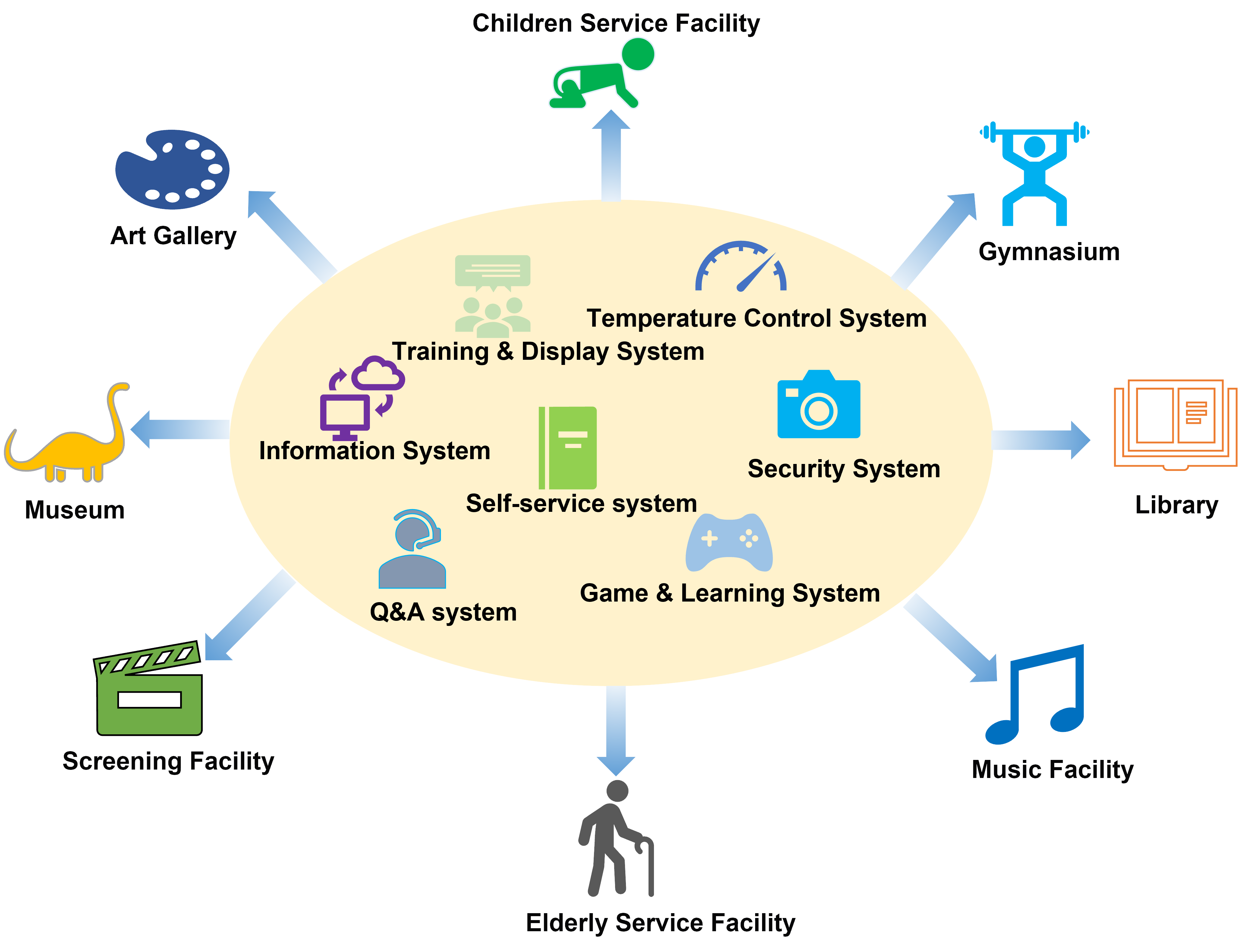}
	\caption{The typical public facilities where smart devices are widely used to provide various basic services.}
	\label{fig_schema}
\end{figure}

To maintain the high availability and efficiency of these public cultural facilities, substantial public expenditure is allocated to the procurement and deployment of smart devices  to enhance management efficiency and user experience \cite{florida2002rise}. These advanced smart devices, such as smart self-service equipment, intelligent climate control devices, and smart security devices, have significantly improved the user experience \cite{chianese2013smartweet}; see Figure \ref{fig_schema}. However, if not properly managed, the key risk performance indicators of these devices, such as reliability, could significantly threaten the sustainable development goals of urban cultural spaces \cite{amato2013talking}. Especially for devices at core central nodes and those frequently used, failure not only poses safety hazards but also weakens their public service functions and can severely impact user experience, increasing public fiscal budget expenditures \cite{chianese2015designing}. Therefore, effectively identifying, analyzing, and predicting the failure risks of these devices is crucial for ensuring the sustainable development of cultural spaces~\cite{ruotsalo2013smartmuseum}.

Given the urgency of intelligent device management in cultural facilities, there is an immediate need for an efficient, intelligent, and convenient approach to integrated management. However, the data generated by various types of devices are heterogeneous and complex \cite{Shi202420769,Xu202468,hu2021distributed}, especially the large volumes of video or image data produced by security systems, along with a significant amount of audio data and textual information. Integrating these diverse and heterogeneous data sources for failure analysis and prevention poses a significant challenge \cite{Bi2023}. The challenges lie in the following aspects.

Firstly, the diversity and complexity of data make traditional data processing methods inadequate. Different types of data have varying formats, structures, and semantics. How to effectively integrate video, audio, and textual data for subsequent analysis and decision-making is an issue that requires immediate resolution \cite{Xu202468,Wu202421072,li2023toward,li2023when}.
Secondly, there is a high demand for real-time processing. The operation of cultural facilities requires real-time monitoring and analysis of device status to identify potential failures and take appropriate measures in a timely manner \cite{Bi2023}. Additionally, data security and privacy protection are critical considerations \cite{wang2019survey,makhdoom2019blockchain,yu2020survey}.  When dealing with audio and video data that involve personal information, how to ensure effective analysis while maintaining data security to prevent data breaches and misuse is a challenge that managers must confront \cite{Xu202468,Thakur202413212,wang2023adversarial}.
Moreover, malfunctions in intelligent devices can cause disruptions to public services, severely degrading the user experience. This is critical in metro public facilities where operational efficiency and passenger satisfaction are paramount \cite{Thakur202413212}. Failures in these systems not only impair service delivery but also undermine the trust and reliability of public service providers. Furthermore, the increasing diversification of specifications and parameters across multi-source devices has led to complexity in identifying the causes of device failures and formulating effective countermeasures. This complexity poses a significant barrier to the integrated management of multi-source intelligent devices, as the intricate nature of these systems can lead to unforeseen interactions and challenges in maintaining seamless operation and service delivery \cite{Bi2023}.

Facing the serious challenges mentioned above, our contributions lie in the following aspects:
\begin{itemize}
	\item We leverage the efficient semantic recognition capabilities of LLMs to meet the diverse intelligent device management needs of users effectively.
	\item Based on the multimodal data processing capabilities of LLMs, we propose an integrated data fusion solution for heterogeneous data from multi-source devices, which efficiently complements each other to achieve fault identification.
	\item With the localized deployment of LLMs, we have achieved data and model localization, fully ensuring the privacy and security of data.
	\item Faced with the increasing diversification of functions and parameters of multi-source intelligent devices, we have wisely proposed a local knowledge extraction based on intelligent device manuals, and provide users with standardized and professional fault-handling suggestions based on this localized expertise.
\end{itemize}

The structure of this article is delineated as follows: The first section delves into the challenges, necessity, and urgency associated with malfunction failures in Internet of Things (IoT) devices for public utilities, as well as the potential benefits of employing LLMs to address these issues. The second section surveys the current industry approaches to fault monitoring and diagnostics utilizing AI technology, pinpointing the unresolved challenges. A solution predicated on LLMs is presented in the third section, which not only addresses the challenges identified in the second section but also proposes an efficient semantic fault prevention solution. The fourth section employs a typical public utility scenario—a library—to conduct simulations and validate the proposed LLM-based fault management solution across different scenarios. The fifth section further discusses the merits of the proposed solution, and the sixth section concludes with a summary and prospects for future work.

\section{Literature Review}

As Artificial Intelligence (AI) and machine learning technologies are becoming increasingly vital,  an increasing number of IoT devices are being managed by various intelligent algorithms. In the domain of fault detection, several articles propose innovative solutions for identifying anomalies and potential failures within various systems. Ribeiro and Mack \cite{Ribiero2024} introduce a multi-vendor device management solution that automates event and fault file collection, which is pivotal for enhancing cybersecurity and NERC compliance among North American utilities, although it may struggle with the seamless integration of diverse vendor technologies. The private cloud-based platform discussed by West and Santitos García \cite{West2021} aims to bolster operational efficiency and grid reliability, potentially through advanced fault detection mechanisms, yet scalability with increasing data volumes remains a concern. Lee and Liu \cite{Lee2023381} present a pluggable module for edge device management, which could encompass fault detection capabilities, although it may not be universally applicable to all edge devices. Liu et al. \cite{Liu20243568} predict power device fatigue failure with a novel approach, which can be seen as a form of fault detection, but it might not cover all failure modes. Enkhbaatar and Yamazaki \cite{Enkhbaatar202222} propose an automatic visual monitoring system for data centers, which is likely to include fault detection to reduce manual checks, but its effectiveness across all hardware anomalies is uncertain. Liu et al. \cite{Liu2021} explore machine learning for substation equipment maintenance, which may involve fault detection to anticipate maintenance needs. Rasheed et al. \cite{Rasheed2023} discuss IoT-enhanced solar device management, which could integrate fault detection for remote surveillance. Sharma et al. \cite{Sharma2024447} utilize deep learning for fault detection and localization in power distribution systems, but the generalizability of their ANN models across different networks is questionable. Balestrieri et al. \cite{Balestrieri2024} present a big data platform for time-series sensor data, which might serve fault detection purposes, though its scalability for increasing sensor data volumes is a concern.

Shifting the focus to fault diagnosis, several articles have delved into diagnosing and addressing the root causes of system malfunctions. Guo et al. \cite{Guo202336452} present DABAC, a smart contract-based access control system for IoT devices, which could assist in diagnosing security breaches within dynamic environments, but its adaptability outside of smart contract platforms is doubtful. Liu et al. \cite{Liu2024} introduce Proteus, a home health monitoring infrastructure that may include diagnostic capabilities for health-related faults, though its suitability for large-scale commercial applications is uncertain. Mavromatis et al. \cite{Mavromatis20201718} propose a software-defined IoT framework that could aid in diagnosing faults in distributed IoT settings, but it may not fully address security concerns. Loukil et al. \cite{Loukil2021} suggest a blockchain-based framework to maintain data privacy, which could involve diagnosing privacy violations. Zhang et al. \cite{Zhang20217333} develop FusionTalk, an IoT-based system for object identification, which might be used for diagnosing identification and tracking issues. Guittoum et al. \cite{Guittoum2023325} propose a multi-agent system for managing IoT cascading failures, which could struggle with diagnosing failures in highly dynamic networks. Ribeiro and Mack \cite{Ribiero202265} summarize a multi-vendor device management system, which might include diagnostic features for NERC compliance, but may not cover all compliance requirements. Ergun et al. \cite{Ergun20233864} propose a dynamic reliability management technique for IoT edge computing, which could face challenges in managing reliability across heterogeneous IoT networks. Angappan et al. \cite{Angappan202589} develop a smart cold storage system with inventory monitoring, which might include diagnostic capabilities for storage conditions, but its optimization for energy consumption is uncertain.

The fault detection component often struggles to handle the challenges related to scalability and integration effectively, especially when dealing with a diverse array of technologies. Moreover, it often falls short of providing comprehensive coverage for all potential failure modes. The fault diagnosis segment, on the other hand, faces doubts regarding the efficiency and adaptability of fault diagnosis in dynamic environments. Its reliance on specific technology platforms, such as smart contracts, may also limit the broad applicability of the solutions. Both segments underscore an urgent need for solutions that are not only robust and adaptable but also comprehensive, capable of navigating the intricacies of fault detection and diagnosis across a spectrum of systems and scales.

Furthermore, a significant shortfall in these segments is the scarcity of work that provides strategies for failure prevention. This is particularly challenging when it comes to managing a multitude of heterogeneous IoT devices, posing a considerable obstacle. The integration of LLMs for IoT failure management offers distinct advantages. These are manifested in semantic interaction and proactive prevention strategies. By semantically integrating diverse and heterogeneous data and combining it with localized knowledge bases specific to particular devices, these models not only efficiently accomplish failure identification and diagnosis but also enable the delivery of effective failure prevention strategies.

The utilization of LLMs in IoT device management represents a paradigm shift. These models excel in their ability to understand and process complex data through advanced natural language processing capabilities. This enables a more nuanced and context-aware approach to failure management, which is crucial in the landscape of IoT where devices operate in varied and often unpredictable conditions. The semantic fusion of multi-source data, augmented by device-specific knowledge, allows for a more precise and swift response to potential failures, thereby enhancing the overall resilience and reliability of IoT ecosystems.

\begin{figure}[h]
	\centering
	\includegraphics[width=\linewidth]{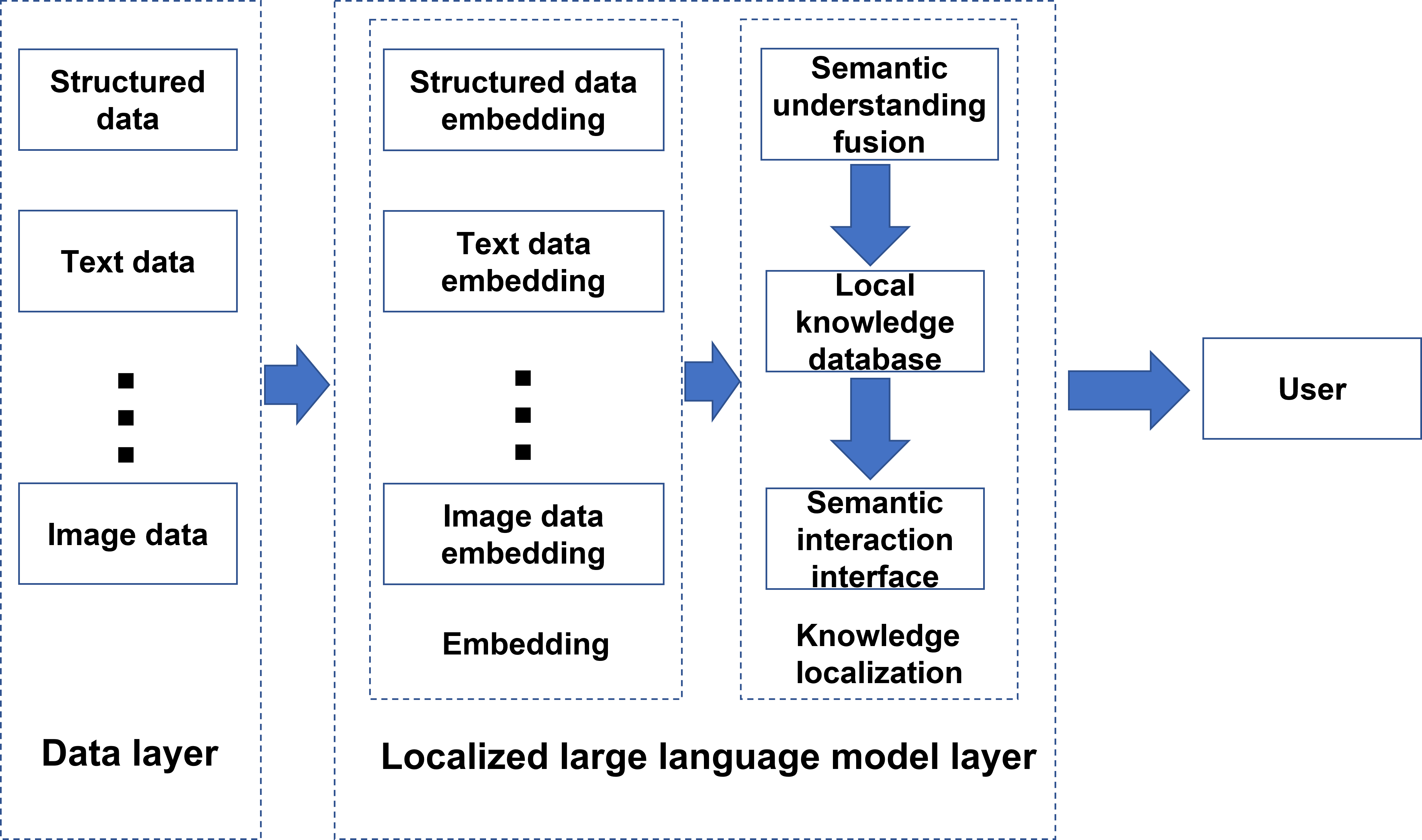}
	\caption{The structure of LLM-based Smart Device Failure Management, where the semantic integration of multi-source heterogeneous data is achieved, and it is combined with the efficient understanding and processing capabilities of LLMs. }
	\label{fig_model}
\end{figure}

\section{Proposed Solution}
Given the challenges previously discussed, we propose an intelligent device management framework based on LLMs that is sustainable and effective. As depicted in Figure \ref{fig_model}, heterogeneous data from multi-source intelligent devices is centrally uploaded and undergoes integration and fusion through a unified semantic approach, serving as an efficient and specialized local knowledge base for LLMs to handle faults and failures.

The framework operates by integrating diverse data sources, such as tables, textual data, image data, and even video data, into a unified semantic representation. This semantic fusion allows for a more comprehensive understanding of the device's operational context, which is crucial for identifying potential failure points. By harnessing the power of LLMs, the framework can interpret and process this semantic information with high efficiency, leading to more accurate and timely detection of device malfunctions. This proactive approach to failure identification and prevention allows for early intervention, reducing the likelihood of device failure and the associated costs and downtime. Moreover, the framework is tailored to meet the specific needs of users, providing them with customized insights and recommendations that enhance their experience with smart devices. This personalization is achieved by analyzing user behavior and preferences, as well as the performance metrics of the devices, to offer targeted suggestions for maintenance and optimization. The end result is an improved user experience, as the devices are more reliable and responsive to user needs.

This framework leverages the semantic integration of multi-source heterogeneous data and combines it with the efficient understanding and processing capabilities of LLMs. It is designed to address the issues of faults and failures in smart devices through customized and efficient failure identification and prevention. By intervening proactively at various stages where devices are at risk of failure, the framework provides constructive suggestions for preemptive maintenance and failure prevention of smart devices. Not only does this framework enable efficient semantic management of smart devices, but it also significantly enhances the quality of service for users' intelligent device usage. Furthermore, it offers constructive recommendations for budgetary decisions in the procurement of equipment for public cultural facilities.

In addition to enhancing user satisfaction, the framework also plays a critical role in supporting decision-making for the procurement of equipment in public cultural facilities. By providing data-driven insights into the performance and maintenance needs of various devices, the framework assists in making informed budgetary decisions. This helps in allocating resources more effectively, ensuring that the equipment is not only cost-efficient but also aligned with the long-term strategic goals of the facility. Overall, the proposed framework is a significant step towards the sustainable and intelligent management of devices, contributing to operational efficiency and user satisfaction in a variety of settings.

\begin{figure}[h]
	\centering
	\includegraphics[width=\linewidth]{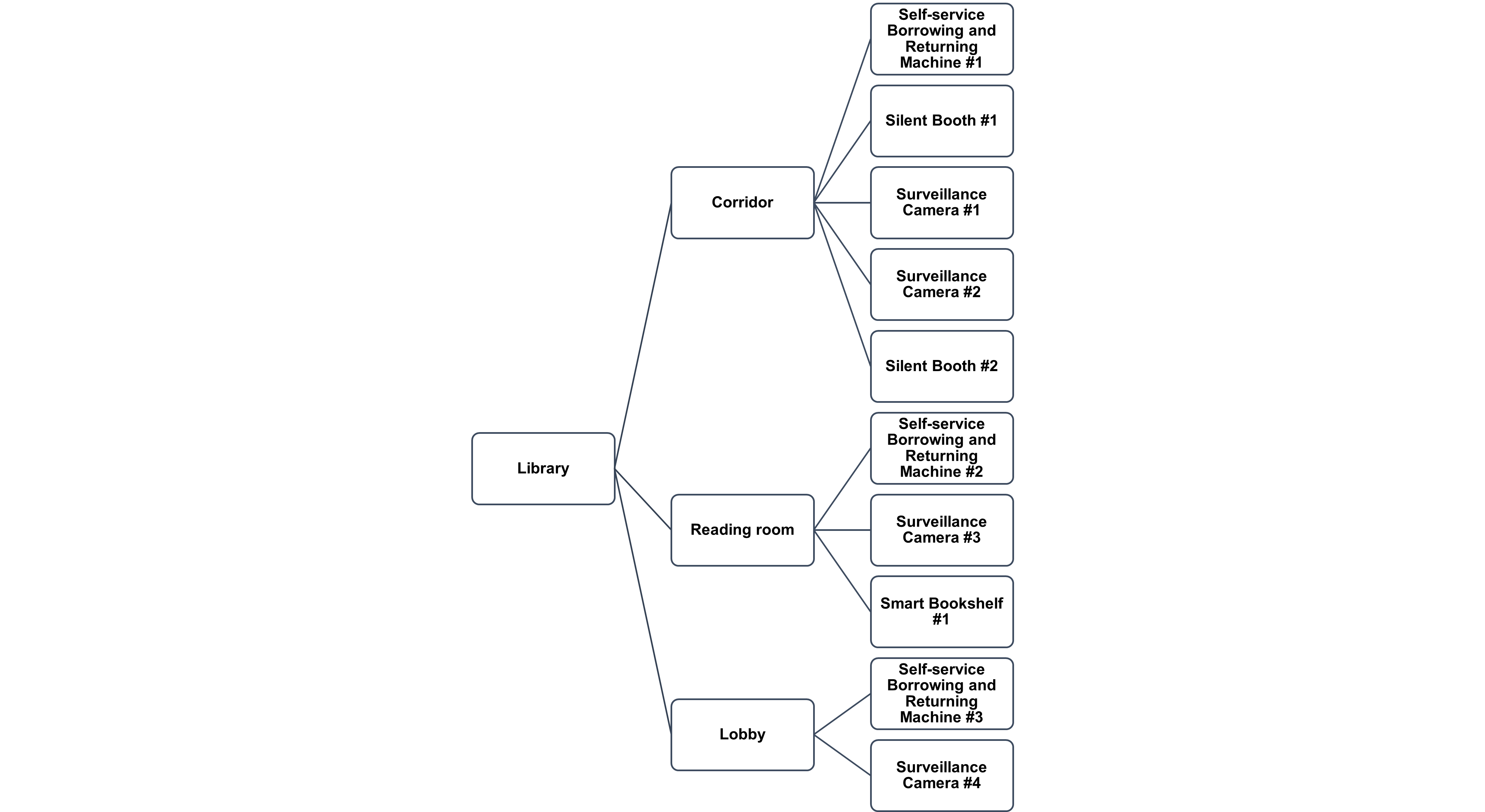}
	\caption{The schema of the considered smart devices deployed in the library for simulation.}
	\label{trial_schema}
\end{figure}

\section{Scene Simulation}
Building upon the concept of scenario simulation, we have chosen the quintessential public cultural facility, the library, as our experimental testbed. Our LLM foundation is based on Llama 3.2-Vision \cite{meta-llama-models}. Our experiment is predicated on three distinct positions within the library: the Corridor, the Reading Room, and the Lobby. Library administrators are keen to implement intelligent semantic management for fault and failure identification and prevention of smart devices in these locations. The smart devices under consideration include the Self-service Borrowing and Returning Machine, Silent Booth, Smart Bookshelf, and Surveillance Camera.

The Self-service Borrowing and Returning Machine, Silent Booth, and Smart Bookshelf primarily generate operational log text data or structured tabular data, which are crucial for monitoring their performance and identifying potential malfunctions. These devices are pivotal to the library's operations, enhancing user experience through seamless services and efficient resource management. On the other hand, the Surveillance Camera, while not directly considered for fault and failure status, plays a significant role in this experiment. It is utilized to test the LLM's ability to process multimodal data. By analyzing images or video footage, the model can monitor the operational status of the other three devices, providing a visual layer of observation that complements the textual and structured data.

This integrated approach allows for a comprehensive understanding of the devices' health and performance. The Self-service Borrowing and Returning Machine, for instance, can be monitored for mechanical failures or software glitches through its log data, while the Silent Booth and Smart Bookshelf can be assessed for user interaction issues or technical malfunctions. The Surveillance Camera's footage can then be analyzed to correlate visual evidence with the data from these devices, offering a holistic view of their operational status. Specifically, traditional front-facing cameras can be replaced with desktop screenshot methods to capture faults. Both approaches have been proven to validate the effectiveness of the solution.

By leveraging the capabilities of LLMs to interpret multimodal data, we can enhance the proactive management of these smart devices. This not only improves the reliability and efficiency of library services but also contributes to a safer and more user-friendly environment. The experiment aims to demonstrate the potential of advanced data analytics in maintaining the optimal functioning of smart devices within public facilities, setting a precedent for the broader application of such technologies in the management of public infrastructure.

\begin{figure}[!h]
	\centering
	\includegraphics[width=8cm]{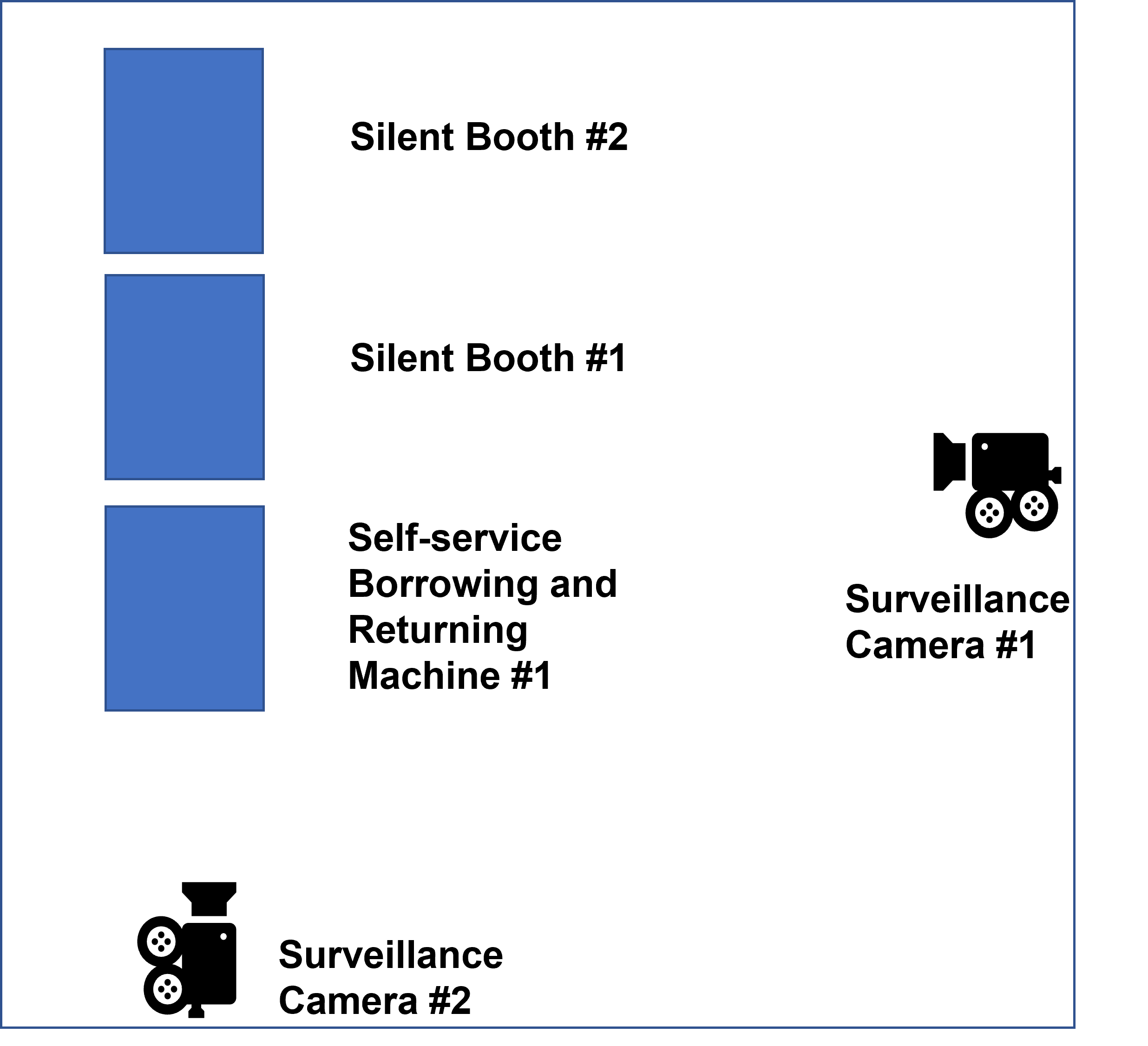}
	\caption{The schematic layout of smart devices in the corridor.}
	\label{corridor}
\end{figure}

\begin{figure}[!h]
	\centering
	\includegraphics[width=8cm]{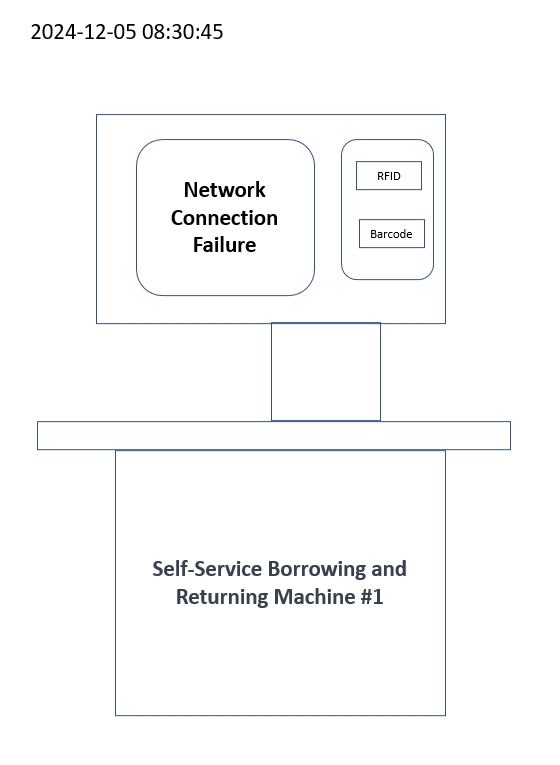}
	\caption{The image of Self-service Borrowing and Returning Machine \#1 taken from the front by Surveillance Camera \#1
		.}
	\label{img_machine_1}
\end{figure}

\begin{figure}[!h]
	\centering
	\includegraphics[width=8cm]{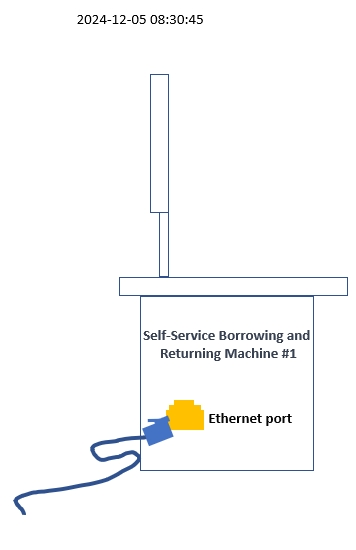}
	\caption{The image of Self-service Borrowing and Returning Machine \#1 taken from the side by Surveillance Camera \#2.}
	\label{img_machine_2}
\end{figure}

\subsection{Test Data}
\begin{figure}[!h]
	\centering
	\includegraphics[width=\linewidth]{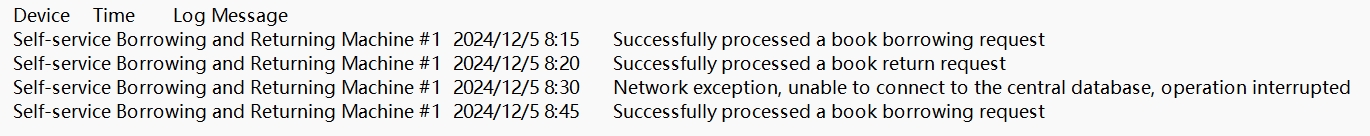}
	\caption{The sample of log information for the Self-service Borrowing and Returning Machine.}
	\label{log_machine}
\end{figure}

The main experimental test scenario is set in a corridor, with the positions of the devices deployed along the corridor shown in Figure \ref{corridor}. It is worth noting that the aforementioned experimental scenario is a simplified demonstration to test the effectiveness of the proposed solution. In actual scenarios, the locations of the various devices can be more flexibly configured according to practical needs. 
We have manually generated log data, based on real scenarios, for the Self-service Borrowing and Returning Machine, as shown in Figure \ref{log_machine}, as well as the synthetic image data from the Surveillance Camera, as shown in Figures \ref{img_machine_1} and \ref{img_machine_2}, to evaluate the supportiveness and efficiency of the proposed solution for handling multi-source heterogeneous data. It is noteworthy that images captured by surveillance cameras often include timestamps. These timestamps, once effectively recognized and semantically organized by LLMs, can be further correlated and integrated with the timestamps from logs. This process allows for the verification of the effectiveness of heterogeneous data fusion from multiple intelligent devices.

\begin{figure}[!h]
	\centering
	\includegraphics[width=\linewidth]{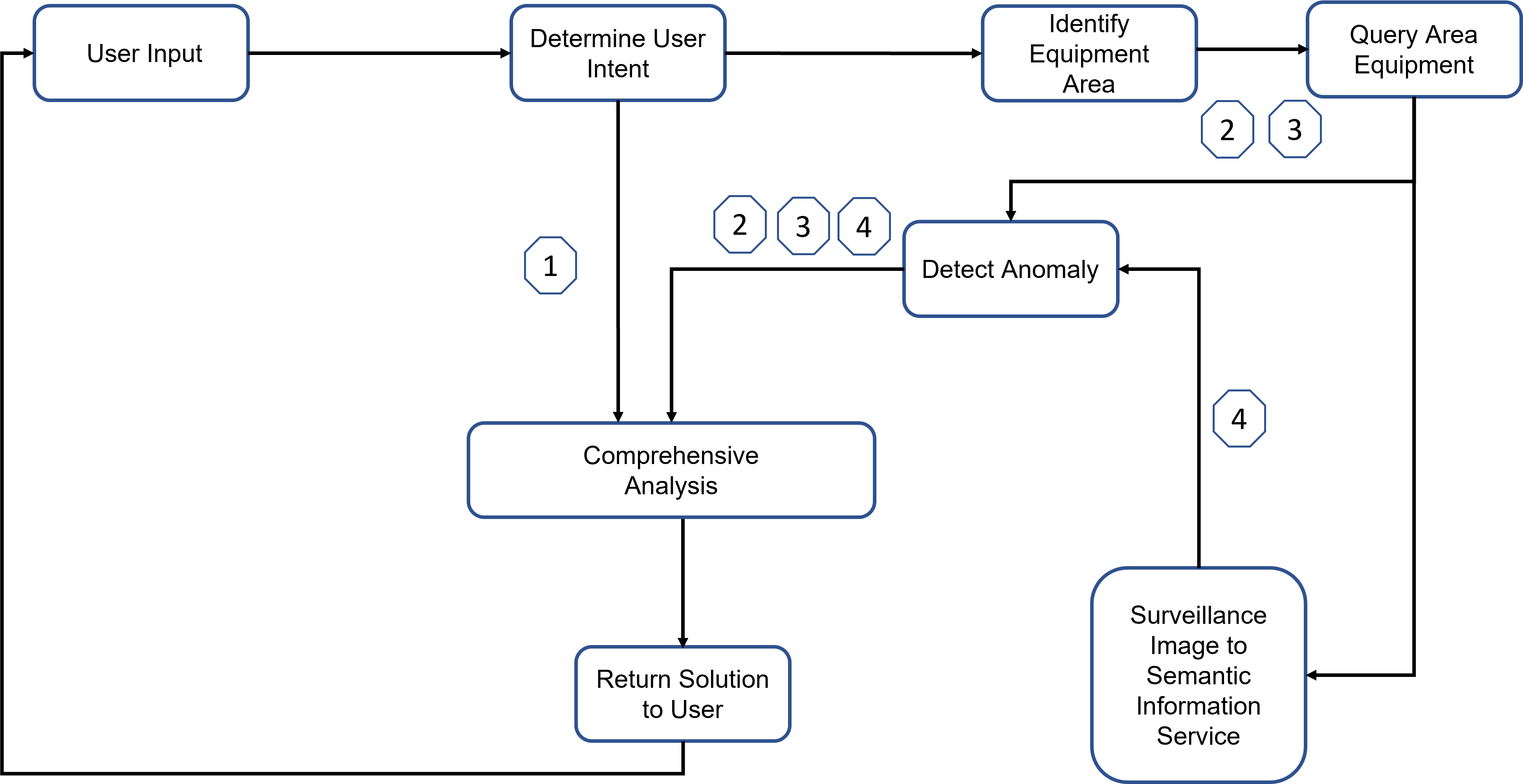}
	\caption{The integrated workflows of four scenarios for failure management, where the key paths corresponding to each scenario are identified with numbers.}
	\label{case_proc}
\end{figure}

\begin{figure}[!h]
	\centering
	\includegraphics[width=10cm]{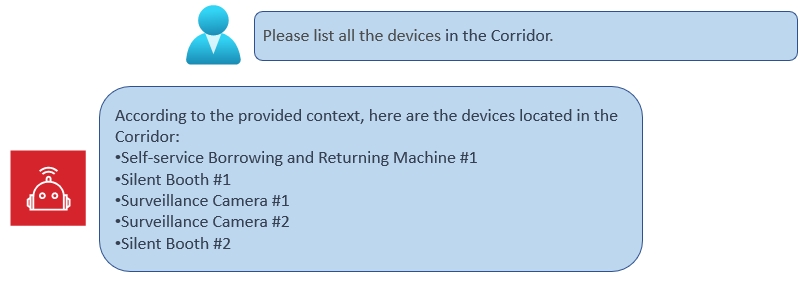}
	\caption{The interactive dialogue for querying the simple statistics of devices deployed in the corridor between the user and the LLM.}
	\label{case1_1}
\end{figure}

\begin{figure}[!h]
	\centering
	\includegraphics[width=10cm]{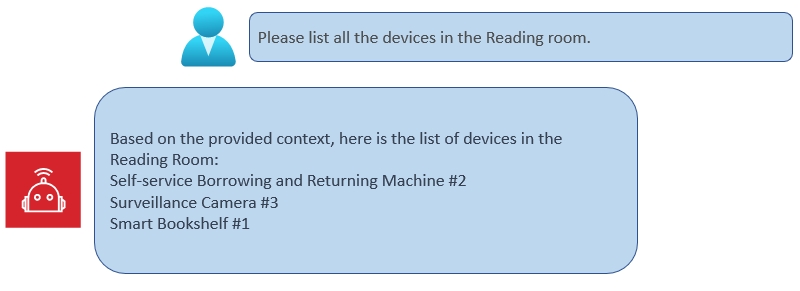}
	\caption{The interactive dialogue for querying the simple statistics of devices deployed in the reading room between the user and the LLM.}
	\label{case1_2}
\end{figure}

\begin{figure}[!h]
	\centering
	\includegraphics[width=10cm]{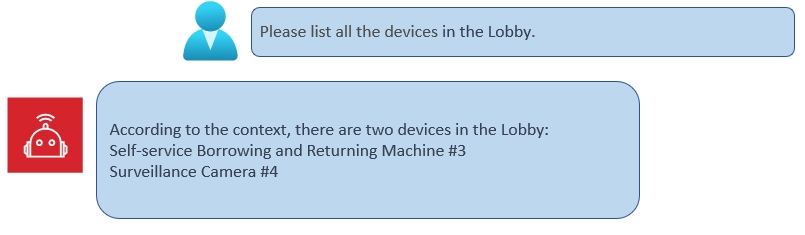}
	\caption{The interactive dialogue for querying the simple statistics of devices deployed in the lobby between the user and the LLM.}
	\label{case1_3}
\end{figure}

\begin{figure}[!h]
	\centering
	\includegraphics[width=10cm]{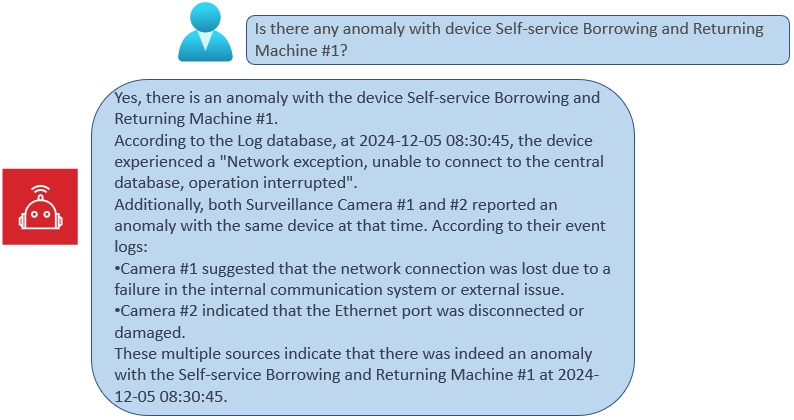}
	\caption{The interactive dialogue for querying the fault and anomaly of devices deployed in the designated area between the user and the LLM.}
	\label{case2}
\end{figure}

\begin{figure}[!h]
	\centering
	\includegraphics[width=10cm]{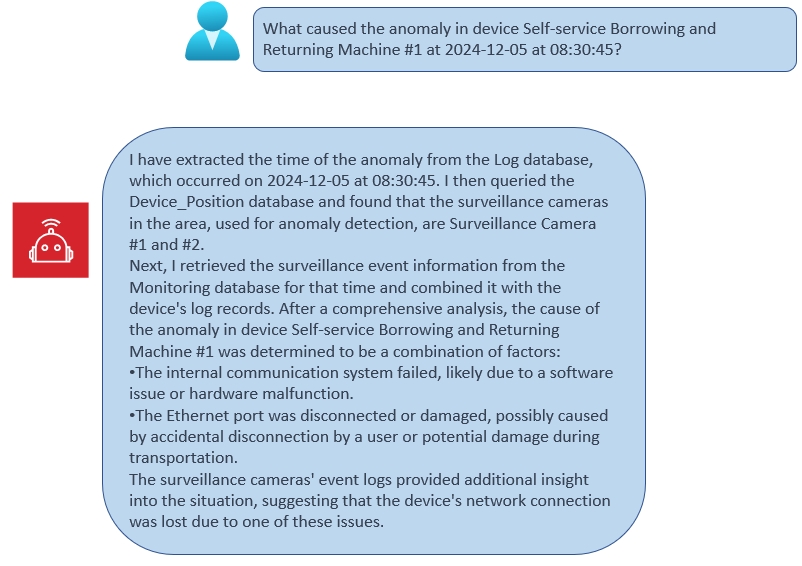}
	\caption{The interactive dialogue for querying the causes resulting in the fault and anomaly of devices deployed in the designated area between the user and the LLM.}
	\label{case3}
\end{figure}

\begin{figure}[!h]
	\centering
	\includegraphics[width=10cm]{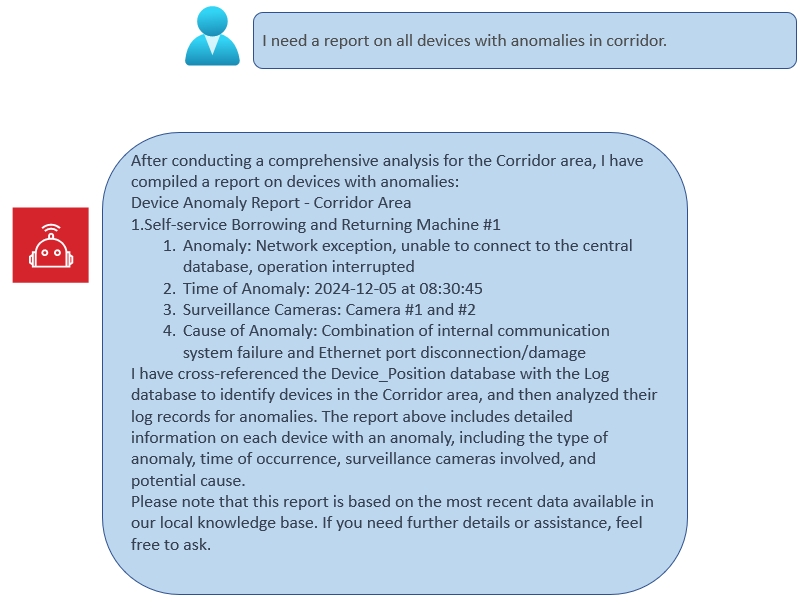}
	\caption{The interactive dialogue for requiring a comprehensive report for the fault and anomaly cases of devices deployed in the designated area between the user and the LLM.}
	\label{case4}
\end{figure}

\begin{figure}[!h]
	\centering
	\includegraphics[angle=90,width=10cm]{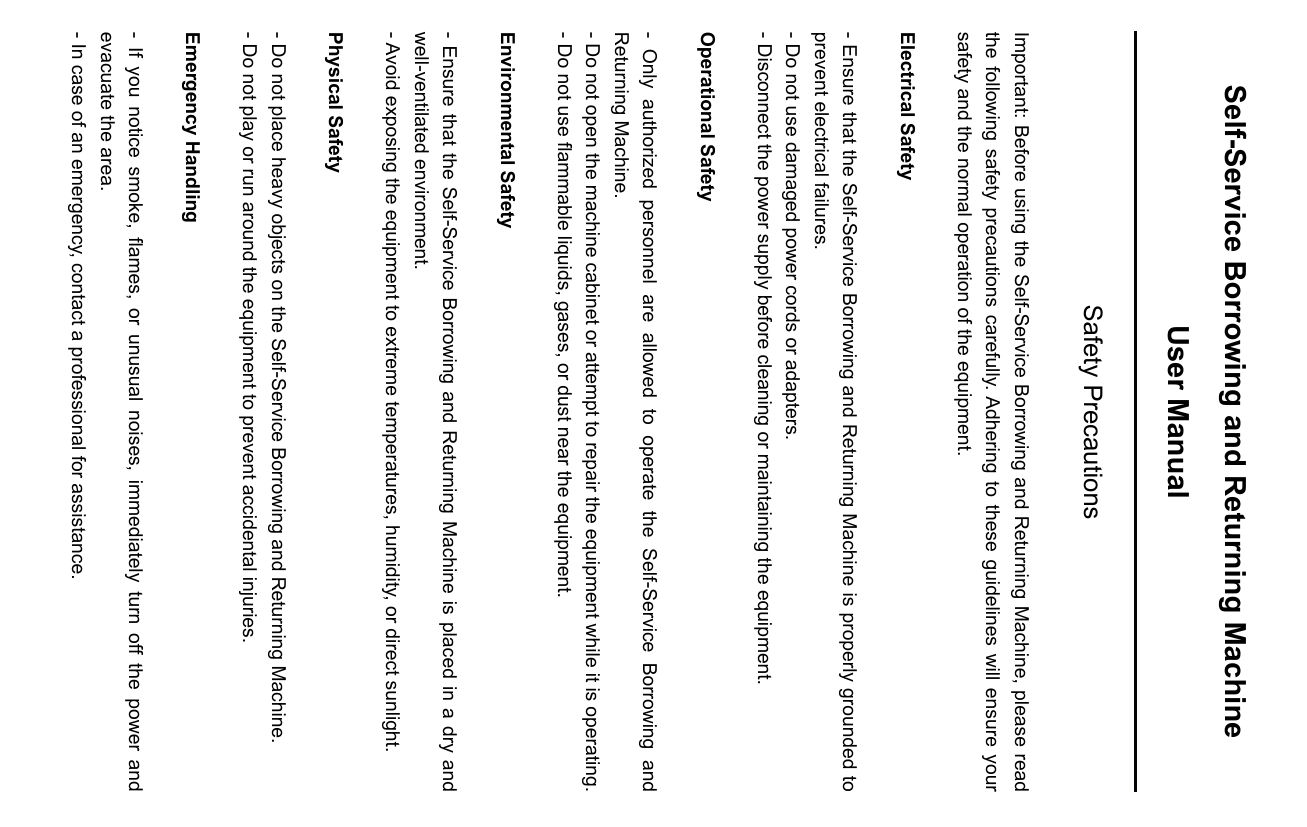}
	\caption{The  sample for the general safety precautions part in the product manual.}
	\label{safety_pre}
\end{figure}

\begin{figure}[!h]
	\centering
	\includegraphics[angle=90,width=10cm]{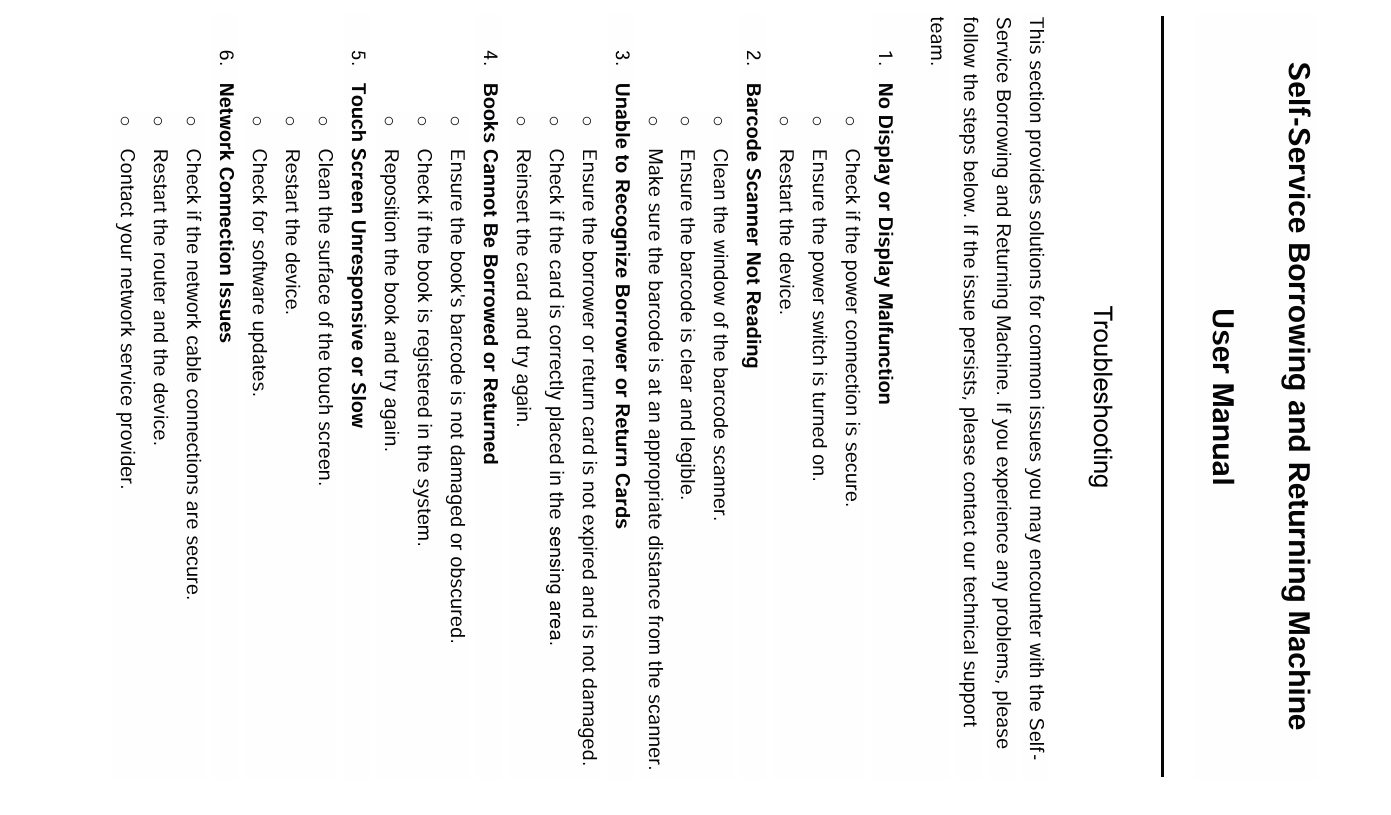}
	\caption{The  sample for the general troubleshooting part in the product manual.}
	\label{trouble_shoot}
\end{figure}

\subsection{Failure Identification and Provenance}
We have contemplated several typical user scenarios to enhance the intelligence of semantic management within the library. The typical work scenarios for failure management include: Inquiry about Device Availability, Fault and Anomaly Queries, Cause of Device Malfunctions, and Comprehensive Risk Assessment and Cause Analysis. We have integrated the workflows of these four scenarios in Figure \ref{case_proc}, where the key paths corresponding to each work scenario are already identified. It is worth noting that in the Path 4 of Figure \ref{case_proc}, we have constructed the capability of the LLM to recognize semantic text from video images into a RESTful local web API service. Similarly, audio files can also leverage web API services (In the experiment, we only considered the semantic transformation of image data, as we are meant to conduct a prototype verification of multi-source heterogeneous data fusion using LLM technology). This could be further imported into the local knowledge base or local database for online querying. Through a semantic approach, we efficiently integrate heterogeneous data generated by multiple smart devices. These scenarios include:

\textbf{Inquiry about Device Availability:} Users may want to know which devices are available in a specific area, corresponding to the Path 1, as labeled in Figure \ref{case_proc}. For instance, a user could ask about the presence of certain smart devices in the Reading Room or the Lobby, seeking to locate them for use. For more detailed information, please refer to Figures \ref{case1_1},\ref{case1_2},\ref{case1_3}.

\textbf{Fault and Anomaly Queries:} Users might inquire about the status of devices within a particular area, specifically whether there are any malfunctions or anomalies, corresponding to the Path 2, as labeled in Figure \ref{case_proc}. A common question could be whether the Self-service Borrowing and Returning Machine in the Corridor is experiencing any issues that would prevent its effective use; see Figure \ref{case2}.

\textbf{Cause of Device Malfunctions:} Users may also seek to understand the reasons behind the malfunctions of specific devices, corresponding to the Path 3, as labeled in Figure \ref{case_proc}. For example, a user might ask why a Silent Booth is not functioning properly, looking for insights into the cause of the problem; see Figure \ref{case3}.

\textbf{Comprehensive Risk Assessment and Cause Analysis:} Generally, users are interested in understanding the broader risks of failure across the library's devices and the causes behind them, corresponding to the Path 4, as labeled in Figure \ref{case_proc}. 
In such complex scenarios, the ability to conduct a comprehensive analysis is built upon the integration of the three aforementioned scenarios, leveraging the sophisticated reasoning capabilities of the LLM and its precision in identifying key processes within intricate workflows. The model must thoroughly understand the true intentions of the users, traverse and search through various paths, and determine the optimal query trajectory to swiftly return accurate information on smart device failures. To achieve this, we have meticulously fine-tuned the LLM with prompt words, enabling it to possess the aforementioned capabilities.
This could involve a request for a comprehensive analysis of the potential failure risks for all smart devices in the library, along with the factors that might be contributing to these issues; see Figure \ref{case4}.

By addressing these user scenarios, we aim to provide a more responsive and informed management system that not only meets the immediate needs of library users but also proactively maintains the efficiency and reliability of the library's smart devices. This approach will lead to a more integrated and user-centric library experience, where the causes of device malfunctions are quickly identified and resolved, minimizing disruption to the library's services.

\subsection{Failure Mitigation and Prevention}

\begin{figure}[h]
	\centering
	\includegraphics[width=10cm]{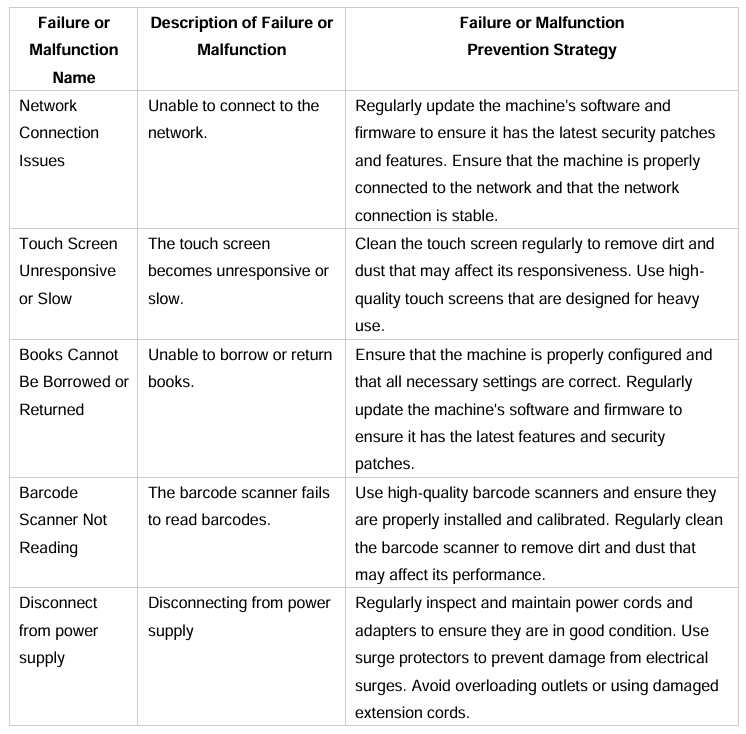}
	\caption{The semantic sample returned by the LLM based on both safety precautions and troubleshooting part in the product manual.}
	\label{failure_prev}
\end{figure}

Previously, we primarily discussed the identification and tracing of faults or failures in smart devices. However, to achieve proactive failure prevention and intervention, a set of strategies for the prevention of faults or failures in smart devices is essential. General large models often possess extensive global knowledge, and to some extent, they can provide users with failure or failure prevention strategies through appropriate prompts in the Prompt. To further supplement the general knowledge of LLMs, extracting failure or failure prevention schemes from the product manuals of smart devices has become a very promising approach for the following reasons:

\begin{enumerate}
	\item Product manuals typically contain product parameters, safety instructions, troubleshooting, and other content, which often implicitly includes specific failure or failure prevention strategies for the product.
	\item Smart device manuals often have copyright protection, but as purchasers of the products, public service institutions might stipulate the usage rights of these documents during the procurement process or contract establishment. Moreover, since the model is deployed locally, the security of the data can be ensured to a certain extent.
\end{enumerate}
Here, we consider two typical parts of the general product manual, i.e., safety instructions and troubleshooting, for the experiment, as shown in Figure \ref{safety_pre} and \ref{trouble_shoot}, respectively.
The sample of extracted information from product manuals is shown in Figure \ref{failure_prev}.
By leveraging the information contained within product manuals, we can enhance the capabilities of LLMs to offer targeted and effective failure prevention strategies. This approach not only leverages the detailed knowledge specific to each device but also ensures compliance with copyright laws while maintaining the privacy and security of the data.
\section{Discussion}
In the rapidly evolving landscape of smart public infrastructure, the integration of advanced technologies plays a pivotal role in enhancing operational efficiency and reliability. Our research has been at the forefront of this technological revolution, focusing on leveraging the power of artificial intelligence, particularly LLMs, to transform the way we manage and maintain intelligent devices in public facilities. Below, we outline our key contributions that not only push the boundaries of current capabilities but also hold the potential to redefine the future of sustainable public infrastructure management.

Our pioneering contribution lies in the innovative application of cutting-edge LLM technology to the realm of public infrastructure. Specifically, we have harnessed this technology for the intelligent analysis and prevention of malfunctions and failures in public facility devices. This innovative approach not only enhances the sustainability of device maintenance and usage but also significantly reduces the budgetary strain on public facilities by optimizing resource allocation and operational efficiency.

We have introduced a novel method of semantically integrating heterogeneous data generated by various intelligent devices through the power of LLMs. This integration is not only a technical feat but a strategic advancement that bolsters the intelligent monitoring of devices, ensuring their longevity and performance. By streamlining data from diverse sources into a cohesive and meaningful format, we support the continuous improvement and sustainability of intelligent device operations.

We have developed and proposed astute strategies for failure prevention and data acquisition that are grounded in practical intelligence. By leveraging localized knowledge bases derived from product manuals, we offer tailored and precise services for the prevention of device malfunctions and failures. This approach is not only proactive but also dynamic, adapting to the unique needs of a wide array of intelligent devices and providing them with the most effective preventive measures.

These contributions represent a significant leap forward in the field of public infrastructure management, where the intelligent use of data and advanced language models can lead to more sustainable, efficient, cost-effective, and resilient systems.

\section{Conclusion and Future Work}

We initially proposed the LLM-based Smart Device Management framework, designed to address the complex challenges of intelligent device management in public facilities. Building upon this foundation, we focused on smart device failure management within a typical representative of public infrastructure— library. We conducted a prototype validation of our proposed framework model in real-world scenarios specific to libraries, demonstrating the effectiveness and feasibility of our approach.

Our work has not only highlighted the versatility and robustness of the proposed LLM-based Smart Device Management framework but also its practical applicability in enhancing the operational efficiency and reliability of intelligent devices in libraries. The prototype validation has fully showcased the advanced and innovative nature of our proposed model, proving its merit in practical settings.

Looking forward, we aim to expand the scope of our framework to encompass a broader range of public facilities beyond libraries. We are committed to further refining our model to handle an even more diverse set of scenarios and to improve its predictive capabilities for device failures. Additionally, we plan to explore the integration of our framework with advanced cybersecurity technologies, such as IoT security and machine learning algorithms designed for threat detection and response. This integration aims to create a more comprehensive and proactive maintenance system that not only enhances the security posture of intelligent devices but also leverages machine learning's capabilities for automated analysis, real-time threat detection, and adaptive security measures. By incorporating these cutting-edge cyber security elements, our framework will be better equipped to address the dynamic and complex challenges of modern public infrastructure, ensuring robust protection against a wide range of potential threats. This will enable public facilities to not only respond to failures but also to anticipate and prevent them, leading to significant cost savings and improved service quality.

%
%
%



\bibliographystyle{elsarticle-num}
\bibliography{sample}
\end{document}